\documentclass[11pt,letterpaper]{uiuc_paper}
\pdfoutput=1
\usepackage[utf8]{inputenc}
\usepackage{amsmath,amssymb,tabularx}
\usepackage{needspace}
\usepackage{xurl}
\definecolor{caseSC}{HTML}{B33A32}
\definecolor{caseResolve}{HTML}{176445}
\newcolumntype{Y}{>{\centering\arraybackslash}X}
\newcommand{\method}{ReSolve}
\newcommand{\scfour}{SC@4}
\newcommand{\sceight}{SC@8}

\renewcommand{\arraystretch}{1.12}
\title{ReSolve: Reusing Candidate Reasoning through Selective Generative Moderation}
\author[1,*]{Bangji Yang}
\author[1,*]{Jiajun Fan}
\author[2,*]{Hongbo Ma}
\author[4]{Xi Zhu}
\author[3]{Weizhi Zhang}
\author[4]{Minghao Guo}
\author[2]{Ye Li}
\author[5]{Hamid Palangi}
\author[1,\textdagger]{Jiaxuan You}
\affiliation[1]{University of Illinois at Urbana-Champaign}
\affiliation[2]{Tsinghua University}
\affiliation[3]{University of Illinois Chicago}
\affiliation[4]{Rutgers University}
\affiliation[5]{Google}
\contribution[*]{Equal contribution}
\contribution[\textdagger]{Corresponding author}
\correspondence{\email{jiaxuan@illinois.edu}}

\abstract{Sampling multiple solutions spends computation on intermediate deductions and unfinished arguments as well as final answers. We introduce ReSolve, a training-free inference procedure that reuses this candidate reasoning through selective generative moderation. An answer-distribution controller invokes a model to examine existing derivations when candidates disagree or lack a parseable answer, then incorporates the generated solution into a bounded loop. Under Hybrid scoring on 130 competition-mathematics problems evaluated with two independently sampled candidate pools, ReSolve obtains 100 and 99 correct answers, compared with 91 and 92 for voting over the same four candidates, with no correct-to-incorrect changes relative to that vote in either pool. Eight-sample self-consistency obtains 94 and 96 correct answers while consuming substantially more tokens; ReSolve uses 46.3\% and 47.2\% fewer tokens in the two evaluations. A controlled ablation removes visible derivations while retaining answer keys, vote counts, and the per-state output-cap rule, reducing accuracy from 100 to 93 correct despite increasing computation. Selective and always-on Uniform moderation both solve 97 problems, while selectivity reduces moderation tokens by approximately 54\% and total pipeline tokens by 6.2\%. These results support candidate reasoning as reusable inference computation. They do not establish an accuracy advantage over additional sampling or a distinct benefit from specialized route instructions.
}
\date{October 2026}
\hypersetup{
  pdftitle={ReSolve: Reusing Candidate Reasoning through Selective Generative Moderation},
  pdfauthor={Bangji Yang, Jiajun Fan, Hongbo Ma, Xi Zhu, Weizhi Zhang, Minghao Guo, Ye Li, Hamid Palangi, Jiaxuan You},
  pdfsubject={Selective generative moderation for mathematical reasoning},
  pdfkeywords={reasoning, inference-time computation, self-consistency, generative moderation}
}

\begin{document}
\maketitle
\section{Introduction}
Sampling several solutions is a common way to allocate additional inference computation to difficult reasoning problems. By the time four candidates have been generated, the model has already spent computation on their decompositions, intermediate identities, and unfinished arguments. Self-consistency aggregates extracted final answers by frequency~\citep{wang2023selfconsistency}, without directly using those arguments in its decision. A partial derivation contributes no vote if it never reaches an answer, and several candidates can repeat the same omission. Before spending more computation on independent attempts, can we make better use of the reasoning already available?

We study selective reuse of these existing derivations. A moderator receives the problem and candidate evidence, checks disputed deductions, and produces a solution that can extend or revise the candidates. A selector must return an existing response. A generative moderator can complete an unfinished argument or construct a corrected answer. Reuse still has a cost: the model must read the evidence and generate another response, and that response can replace a correct answer with an incorrect one. The allocation decision is therefore both what evidence to reuse and when another call is worth making.

\method{} implements this idea as selective generative moderation. It samples four initial solutions, groups their parseable answers, and bypasses moderation when the current pool has a single answer group. Otherwise, the answer distribution selects an instruction to inspect a dominant answer, compare two groups, reconcile fragmented evidence, or complete unfinished attempts. The resulting response enters the pool, and an explicit acceptance rule either returns a solution or continues within a bounded number of calls. The same existing model proposes and moderates; no reference answers are available during inference.

The key distinction is between \emph{reusing evidence} and \emph{requesting another independent attempt}. We use shared candidates and count all input and output tokens. Under the primary Hybrid score, two independently sampled candidate pools on the same 130 competition-mathematics problems yield 100 and 99 correct answers for \method{}, versus 91 and 92 for voting over the same four candidates. Moderation corrects nine and seven errors, respectively, with no correct-to-incorrect changes relative to that vote in either pool. Relative to eight-sample self-consistency, the observed counts are 100 versus 94 and 99 versus 96, with 46--47\% fewer total tokens. A secondary paired analysis that groups the two evaluations by problem supports improvement over four-candidate voting, while the contrast with eight-candidate voting remains statistically inconclusive.

We then isolate what information and control decisions produce these observations. Removing visible candidate derivations lowers the first evaluation from 100 to 93 correct answers, although the answer-only moderator makes more calls and generates more tokens. Keeping a generic verification instruction and changing only whether agreement states are checked leaves accuracy at 97 correct while more than doubling moderation tokens. By contrast, route-specific instructions do not establish an advantage over a generic Uniform moderator. These findings place the empirical emphasis on \emph{what evidence to reuse} and \emph{when to invoke another model call}.

Our contributions are threefold.
\begin{enumerate}
\item \textbf{Candidate reasoning as reusable computation.} We formulate existing derivations as evidence for further inference and distinguish generative recovery from fixed-pool answer selection.
\item \textbf{Selective generative moderation.} We instantiate this formulation with an explicit, bounded procedure that reads candidate evidence when answers disagree or are unavailable. Its token accounting includes both evidence access and generation.
\item \textbf{Controlled evidence for reuse and selectivity.} We evaluate two candidate pools, remove visible derivations, and compare selective with always-on invocation. Recovery and preservation analyses expose both what moderation repairs and what it can lose.
\end{enumerate}
The empirical scope is mathematical reasoning with a 4B backbone. Fresh candidate draws test sensitivity to the sampled pool; they do not by themselves establish generalization to new tasks or rule out benchmark familiarity.

\section{ReSolve: selective generative moderation}
\label{sec:method}
\subsection{From candidate answers to candidate evidence}
For a problem $x$, a proposer $p_\phi$ generates $K$ candidate solutions $C_0=\{y_1,\ldots,y_K\}$. Let $a(y)$ extract and normalize a final answer, with $a(y)=\bot$ when extraction fails. The answer histogram is
\begin{equation}
 h_C(v)=\sum_{y\in C}\mathbb{1}[a(y)=v],\qquad
 n_C=\sum_{v\ne\bot}h_C(v).
\end{equation}
Self-consistency returns a representative of the most frequent valid answer. \method{} additionally retains the reasoning behind those answers. Its evidence constructor $e(x,C)$ packages the problem, representative arguments, and the available answer-distribution information for a moderator $q_\theta$.

\begin{figure}[htbp]
\centering
\includegraphics[width=\linewidth]{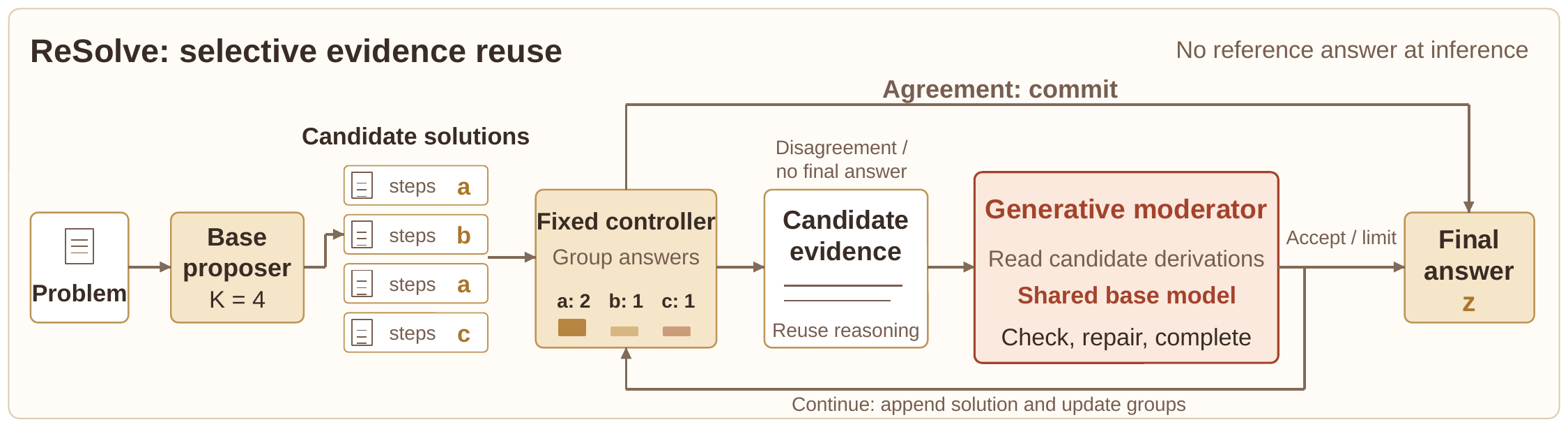}
\caption{\textbf{Selective reuse of candidate reasoning.} Four sampled solutions provide both an answer distribution and visible derivations. Agreement permits a direct decision; otherwise, a generative moderator checks, repairs, or completes candidate reasoning. New solutions update the pool until acceptance or the call limit. Proposer and moderator share existing model weights, and reference answers are absent at inference.}
\label{fig:framework}
\end{figure}

The distinction between selecting and generating is precise. With an evaluation-only correctness function $g_x$, the fixed-pool selection oracle is
\begin{equation}
 O_K(x)=\max_{y\in C_0} g_x(y).
\end{equation}
Any procedure restricted to returning an element of $C_0$ is bounded by $O_K$. A moderator generates $z\sim q_\theta(\cdot\mid e(x,C))$ and can succeed when $O_K=0$. Over a dataset, the difference between moderation and oracle accuracy decomposes as
\begin{equation}
 \sum_x\!\big(g_x(z_x)-O_K(x)\big)=N_{\mathrm{recover}}-N_{\mathrm{lose}},
 \label{eq:oracle}
\end{equation}
where $z_x$ denotes the system's final returned response, either an existing candidate or a new generation. A recovery solves an oracle-unsolved problem and a loss fails on an oracle-solvable one. This accounting identity makes both effects visible. The oracle is an evaluation control with reference-answer access; it is not a deployable system or an upper bound on additional generation.

\subsection{Selective intervention and bounded feedback}
\label{sec:controller}
The controller partitions the current histogram into five states (Table~\ref{tab:states}). Agreement permits a direct decision. For disagreement, the moderator checks a majority argument, compares two answer groups, or reconciles more fragmented evidence. If no final answer can be extracted, unfinished attempts are supplied as weak evidence for repair. None of these inference operations receives the reference answer or correctness feedback.

\begin{table}[htbp]
\centering\small
\caption{Controller states, evaluated in the displayed order. Shares use the number of \emph{parseable} answers. In particular, agreement can occur with fewer than $K$ parseable candidates. We use $\tau=0.5$.}
\label{tab:states}
\begin{tabularx}{\linewidth}{c c Y}
\toprule
State & Histogram condition & Intervention \\
\midrule
No final answer & $n_C=0$ & Complete or repair unfinished attempts \\
Agreement & One valid answer group & Commit without a new call \\
Dominant answer & Top share $>\tau$ & Audit the majority and competing evidence \\
Two groups & Two groups, no dominant answer & Adjudicate the competing arguments \\
Fragmented & More than two, no dominant answer & Reconcile useful deductions \\
\bottomrule
\end{tabularx}

\end{table}

The evidence view retains the first representative of each answer group and its multiplicity, ordered by frequency with candidate-order tie breaking. A later derivation supporting an existing key increases its count without replacing that representative. A new key can introduce a new representative. Rendering prefers the post-\texttt{</think>} derivation when nonempty and otherwise uses the original response, within the same evidence limits across compared methods. No-answer repair instead reads a bounded prefix of the current attempts (Appendix~\ref{app:implementation}).

\paragraph{Feedback updates answer support and selected evidence.}
Generated responses join the pool, but later calls do not necessarily see every new derivation. For example, consider a pool with answer counts $\{a:3,b:1\}$. If moderation returns $b$, the pool becomes $\{a:3,b:2\}$ and still calls for a majority audit. A second parseable moderator response of $b$ meets the stability criterion and permits an overturn; a response of $a$ instead confirms the leading answer. In two-group or fragmented states, selecting an existing answer permits immediate acceptance, while a new answer requires continued processing. These are deterministic acceptance conditions on answer keys, not certificates of mathematical correctness. An unparseable response does not reset that sequence; its text is eligible for no-answer repair but is not an answer-group representative.

After each moderation call, its answer is appended to the pool and the histogram is recomputed. State-specific acceptance and a stability check determine whether to return a solution. The procedure allows at most $R=3$ moderation calls and uses a stability patience of two. At the round limit it applies the deterministic answer-selection fallback. The full acceptance rules and extraction conventions are given in Appendix~\ref{app:implementation}. This separation between \emph{whether to invoke}, \emph{what to ask}, and \emph{when to stop} allows each component to be examined experimentally.

\subsection{A fixed model and an explicit inference policy}
The proposer and moderator share the same Qwen3-4B-Thinking-2507 weights~\citep{yang2025qwen3}. No parameter update is required. The moderator receives the problem and the rendered candidate evidence; its generated response enters the same answer-extraction and grouping pipeline as the original proposals. Reference answers and evaluator feedback are absent from this process.

\subsection{Accounting for inference computation}
For each problem we count all proposal and moderation input and output tokens:
\begin{equation}
 T(x)=\sum_{i=1}^{K}\big(T^{\mathrm{in}}_i+T^{\mathrm{out}}_i\big)
 +\sum_{t=1}^{R_x}\big(T^{\mathrm{in}}_{m,t}+T^{\mathrm{out}}_{m,t}\big).
\end{equation}
The first sum is shared by four-candidate voting and moderation: both have already generated the same solutions. The second sum measures the additional cost of reading and extending that evidence, and vanishes when the controller bypasses moderation. Eight-sample self-consistency instead generates four more independent solutions. We report the full token consumption of each choice, including reading candidate evidence. Token counts characterize inference workload; they do not assert equivalent latency or FLOPs for arbitrary model and serving configurations.

\section{Experiments}
\label{sec:experiments}
\subsection{Evaluation design}
\paragraph{Two candidate pools on the same problems.}
The principal evaluation contains AIME24, AIME25, and HMMT25 (30 problems each), plus CMIMC25 (40), for 130 distinct problems. F2 and F3 denote two independently sampled candidate pools for these same problems. Each evaluation generates eight Qwen3-4B solutions per problem. \scfour{}, Oracle@4, \method{}, and Uniform use the first four; \sceight{} and Oracle@8 use all eight. The additional four candidates are hidden from the moderators. F3 keeps the questions, frozen inference rules, and grading code unchanged. Its batched serving implementation was checked for equivalent per-problem controller behavior. These are repetitions of candidate sampling on previously studied questions, not additional independent test problems.

\paragraph{Controls and generation.}
\emph{Uniform} uses a general verification instruction with the same visible evidence, invocation gate, acceptance conditions, and stopping controller as \method{}. \emph{Answer-only} removes the visible derivations while retaining answer keys and counts. \emph{Direct32} solves independently without candidate evidence or voting fallback. Proposal temperature/top-$p$ are 0.7/0.9; moderation and Direct32 use 0.1/1.0. Calls share a 32,768-token context window, with output capacity equal to the remaining context after the prompt and a 32-token margin. Answer-only uses the capacity computed from its state's Full rendering to isolate evidence removal. We retain historical same-pool studies H1/D1 for supporting mechanism analyses, with their separate 16K output budget documented in the appendix.

\paragraph{Scoring and paired analysis.}
We use the prespecified Hybrid score, which accepts an answer when either the frozen rule grader or CompassVerifier-3B (CV) accepts it~\citep{opencompass2025compassverifier}. CV is a sensitivity analysis. A scoring audit identified an equivalent-form false negative in the CV-only SC@4 and SC@8 scores, reinforcing the need to distinguish the two views (Section~\ref{sec:grading}). Within each candidate pool, we report exact two-sided McNemar tests. Holm correction covers the two prespecified contrasts, \method{} versus \sceight{} and versus Uniform. The same-pool \scfour{} comparison is a secondary analysis. Across F2 and F3, the problem is the statistical unit: the reported sign-flip analysis groups its two paired outcomes rather than treating 260 outcomes as independent problems. Appendix~\ref{app:statistics} separates these tests and their scope.

\subsection{Reusing four candidates versus sampling eight}
\label{sec:mainresults}
\begin{table}[htbp]
\centering\small
\caption{Hybrid results on two independent candidate pools for the same 130 problems. ReSolve and Uniform access only the first four candidates. Oracle rows use evaluation reference answers. The evaluations are repeated observations of the same questions.}
\label{tab:main}
\begin{tabular}{ccccc}
\toprule
Method & F2 correct & F2 (\%) & F3 correct & F3 (\%) \\
\midrule
SC@4 & 91 & 70.00 & 92 & 70.77 \\
SC@8 & 94 & 72.31 & 96 & 73.85 \\
Direct32 & 82 & 63.08 & 80 & 61.54 \\
Uniform & 97 & 74.62 & 99 & 76.15 \\
\method{} & 100 & 76.92 & 99 & 76.15 \\
\midrule
Oracle@4 & 99 & 76.15 & 100 & 76.92 \\
Oracle@8 & 100 & 76.92 & 108 & 83.08 \\
\bottomrule
\end{tabular}

\end{table}

\paragraph{Improvement over the same candidate vote.}
Table~\ref{tab:main} reports Hybrid results for both independently sampled pools. \method{} improves \scfour{} from 91 to 100 correct in F2 and from 92 to 99 in F3: gains of 6.92 and 5.38 percentage points. The paired changes are nine fixes with no breaks and seven fixes with no breaks, respectively. In F2, 74 of the 91 SC@4-correct problems bypass moderation; all 17 that are actually moderated remain correct. Across both pools, 16 fixes and no breaks give a mean gain of 6.15 percentage points. These are repeated observations of 130 problems, not 260 distinct questions.

The secondary paired analysis gives within-pool $p=0.0039$ and $p=0.0156$, and combined problem-level $p=0.0003$ (Appendix~\ref{app:statistics}). F2 moderation adds only 6.4\% to the token cost of the shared four proposals. This comparison asks what reading and extending already-generated reasoning adds beyond its final-answer vote.

\paragraph{A different allocation of the inference budget.}
\sceight{} solves 94 problems in F2 and 96 in F3, compared with 100 and 99 for \method{}. The F2 totals remain 100 versus 94 under the rule grader alone (Table~\ref{tab:scorers}). F2 has seven wins and one loss ($p=0.0703$, Holm $p=0.1406$); F3 has seven wins and four losses ($p=0.5488$, Holm $p=1$). The combined gain is 3.46 percentage points with problem-level $p=0.093$. The observed accuracy ordering repeats, but its margin narrows and does not establish superiority over additional sampling. Nor does a nonsignificant difference establish equivalence.

\begin{figure}[htbp]
\centering
\includegraphics[width=\linewidth]{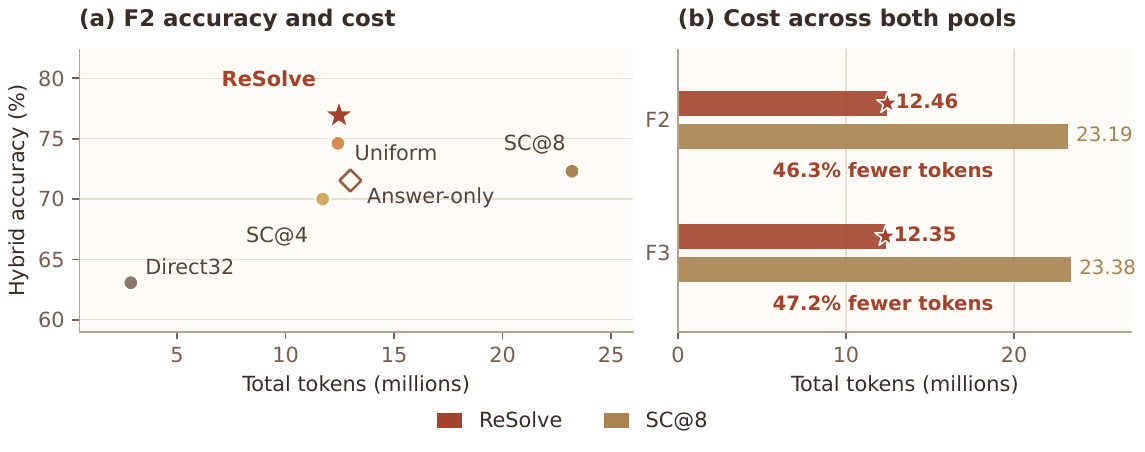}
\caption{\textbf{Accuracy and token cost with Hybrid scoring.} (a) F2 controls, including the answer-only intervention; the star marks \method{}. (b) Whole-pipeline token cost for \method{} and SC@8 on both independent candidate pools. Each method's total includes the proposals it uses and all moderation input/output tokens.}
\label{fig:cost}
\end{figure}

The token comparison is more consistent. \method{} uses 12.46M versus 23.19M tokens in F2 and 12.35M versus 23.38M in F3: reductions of 46.3\% and 47.2\% (Figure~\ref{fig:cost}). For F2, the shared four proposals consume 11.711M tokens; moderation adds 0.753M, including 0.195M input tokens for reading evidence and 0.558M output tokens. The next four independent proposals consume 11.483M. The cost advantage therefore includes the price of accessing previous reasoning rather than treating evidence as free input. F3 makes 47 moderation calls, compared with 51 in F2, with no call reaching its output cap. SC@8 adds samples on every problem, whereas moderation is selective; this comparison changes both which problems receive computation and what computation they receive.

\paragraph{What the instruction controls establish.}
Uniform solves 97 problems in F2 and 99 in F3. The \method{}--Uniform paired outcomes are 4/1 and 2/2 wins/losses; combined problem-level $p=0.51$. Specialized route instructions therefore do not establish an improvement over generic candidate-conditioned verification. Direct32 solves 82 and 80 problems, but uses a different budget; its lower F2 cost of 2.88M tokens prevents interpreting its accuracy gap as a matched-cost comparison. The central evidence concerns reuse versus voting and the observed tradeoff against further sampling.

\subsection{Does the visible candidate reasoning matter?}
\label{sec:mechanism}
\begin{table}[htbp]
\centering\small
\caption{F2 controls under Hybrid. Moderation tokens include inputs and outputs; all rows share an additional 11.711M proposal tokens. Full versus Answer-only removes derivations under the same per-state output-cap rule. Selective versus always-on Uniform adds agreement checks, shared across the two Uniform variants.}
\label{tab:evidenceselectivity}
\begin{tabular}{cccc}
\toprule
F2 configuration & Correct / 130 & Calls & Moderation (M) \\
\midrule
\method{} (Full) & 100 & 51 & 0.753 \\
Answer-only & 93 & 65 & 1.276 \\
\midrule
Uniform loop, selective & 97 & 50 & 0.706 \\
Uniform loop, always-on & 97 & 136 & 1.522 \\
Uniform one-call, selective & 97 & 44 & 0.619 \\
Uniform one-call, always-on & 97 & 130 & 1.435 \\
\bottomrule
\end{tabular}

\end{table}

\paragraph{A controlled removal of derivations.}
Answer-only retains the problem, group headers, extracted answers, vote counts, route instructions, decoding settings, and controller. It deletes the rendered derivations from both original and subsequently generated candidates. Each call's cap is computed from the Full rendering of that same state. Shorter input consequently does not grant extra output capacity. All 44 initially triggered problems match the original Full prompt and cap before removal; the 86 agreement outputs are reused unchanged. The intervention concerns visible written solutions, not hidden thinking traces.

Full solves 100 problems versus 93 for Answer-only, with seven fixes and zero breaks ($p=0.0156$). Four fixes have parseable candidate answers: two dominant-answer states, one two-group state, and one fragmented state. The remaining three have no extracted initial answer. The benefit therefore extends beyond completing unfinished attempts. In the four disagreement cases, the answer-only moderator ends on an existing incorrect answer group. Full evidence can support either preserving the correct group or overturning an incorrect majority.

Answer-only makes 65 calls versus 51 for Full and generates 1.257M versus 0.558M moderator tokens. Its total pipeline cost is approximately 12.99M rather than 12.46M, despite reducing moderation input from 0.195M to 0.019M tokens. Twenty answer-only calls reach the output cap, compared with one Full call. These observations are consistent with existing derivations reducing the work needed to reach a solution within the tested budget. They do not prove the same gain under a different cap policy. In particular, allowing the shorter answer-only prompt to spend its saved context on longer generation remains a separate control.

\subsection{Does selective invocation save computation?}
\label{sec:selectivity}
The final-protocol F2 control adds one Uniform verification call to each of the 86 agreement problems and reuses the existing outputs for the triggered problems. A parseable verification answer replaces the previous response; otherwise the original response is retained. This isolates the effect of checking the states normally skipped by the gate.

Selective and always-on Uniform both solve 97 problems, with no correctness changes across the 86 added checks. Eighty-five checks return parseable answers and one changes the extracted answer without changing correctness. The closed-loop comparison uses 50 versus 136 total calls and 0.706M versus 1.522M moderation tokens. Selectivity saves approximately 54\% of moderation tokens; the total pipeline reduction is approximately 6.2\% because the shared proposals dominate cost. A one-call version reaches the same 97 correct with 44 versus 130 calls and 0.619M versus 1.435M moderation tokens (Table~\ref{tab:evidenceselectivity}).

This control supports the agreement bypass on this pool, not the proposition that agreement guarantees correctness. Twelve of the 86 agreement states are wrong, and additional Uniform checking does not repair them. Selectivity avoids work that did not improve outcomes in the observed control. A different verifier or a targeted disagreement-inducing intervention could behave differently; such interventions are not evaluated here.

\subsection{Where gains and preservation failures occur}
\label{sec:routes}
\begin{table}[htbp]
\centering\small
\caption{F2 fixes/breaks by initial state under Hybrid. Comparators isolate same-pool voting, additional proposals, and the moderation instruction. Small strata are descriptive.}
\label{tab:routes}
\begin{tabular}{ccccc}
\toprule
F2 initial state & $N$ & vs. SC@4 & vs. SC@8 & vs. Uniform \\
\midrule
Agreement & 86 & 0/0 & 0/0 & 0/0 \\
Dominant answer & 18 & 2/0 & 2/0 & 3/0 \\
Two groups & 8 & 1/0 & 0/0 & 0/0 \\
Fragmented & 12 & 3/0 & 2/1 & 1/1 \\
No answer & 6 & 3/0 & 3/0 & 0/0 \\
\bottomrule
\end{tabular}

\end{table}

Table~\ref{tab:routes} locates the gains by initial controller state. Under Hybrid scoring, F2's nine fixes over \scfour{} arise from dominant-answer (two), two-group (one), fragmented (three), and no-answer states (three), with no breaks. Against \sceight{}, the net improvements are two dominant-answer, one fragmented, and three no-answer problems. Against Uniform, all of the net gain comes from dominant-answer states. This decomposition identifies where improvements occur; it does not establish a separate advantage for each instruction.

Fixed-pool oracle accounting exposes a complementary failure mode. In F2, \method{} recovers three no-answer problems and loses two oracle-solvable problems. These recoveries show completion of unfinished attempts, not repair of four completed but incorrect answers. In F3, it recovers two and loses three. The corresponding totals are 100 versus Oracle@4's 99, and 99 versus Oracle@4's 100. Generative recovery is possible in both pools, but exceeding the oracle in aggregate is not a stable result. In F3, three of the four losses to \sceight{} occur in no-answer states, where the additional proposals can supply a valid answer unavailable to the moderator. Generating more candidates and completing existing reasoning address different failure modes.

Historical controls are consistent with this interpretation. On H1, \method{} solves 127/160 compared with 119 for \scfour{}, with four recoveries and three preservation losses relative to the Hybrid oracle. In D1, one Uniform call with evidence solves 126 versus 119 for a fresh solve on the same triggered problems under the same 16K output cap. The earlier comparison also changes prompt wording; the F2 answer-only intervention more directly isolates visible derivations. H1's Uniform and operation-permutation controls remain within two correct answers of \method{} (Appendix~\ref{app:historical}).

\subsection{A case of completing candidate evidence}
\label{sec:cases}
\begin{table}[htbp]
\centering
\caption{\textbf{Completing candidate reasoning on CMIMC 2025.} The initial pool has no extractable final answer, but contains an identity that supports a correct completion. All descriptions are edited summaries, not verbatim model quotations.}
\label{tab:case}
\small
\setlength{\tabcolsep}{7pt}
\renewcommand{\arraystretch}{1.08}
\begin{tabularx}{\linewidth}{@{}Y|Y@{}}
\toprule
\multicolumn{2}{@{}>{\centering\arraybackslash}p{\linewidth}@{}}{\textbf{Example: a quartic root (CMIMC25, cache index 7).}
\textbf{Problem:} Let $P(x)=x^4+20x^3+29x^2-666x+2025>0$ for real $x$.
A first-quadrant root has the form $r=(a+bi+\sqrt{c+di})/2$, with integer $a,b,c,d$.
Find $a+b+c+d$. \quad \textbf{Reference answer: 322.}}\\
\midrule
\textcolor{caseSC}{\textbf{Initial candidate pool / SC@4}} & \textcolor{caseResolve}{\textbf{\method{}}}\\
\midrule
\begin{minipage}[t]{\linewidth}
\centering
\color{caseSC}\itshape
All four attempts lack an extractable final answer, so answer voting has no valid answer to select. However, the fourth attempt already contains
\[
\color{black}P(x)=(x^2+10x-36)^2+(x+27)^2.
\]
\textcolor{black}{This useful intermediate identity remains visible in the moderator's actual input.}
\end{minipage}
&
\begin{minipage}[t]{\linewidth}
\centering
\color{caseResolve}
Checks the decomposition and completes the complex quadratic:
\[
x^2+(10-i)x-(36+27i)=0.
\]
The discriminant is $243+88i$, giving
\[
r=\frac{-10+i+\sqrt{243+88i}}{2}.
\]
Thus $-10+1+243+88=322$.
\end{minipage}\\
\midrule
\textcolor{caseSC}{\textbf{Extracted answers:} $[\bot,\bot,\bot,\bot]$} & \textcolor{caseResolve}{\textbf{One moderation call.}\quad Answer: $\boxed{322}$\;\checkmark}\\
\bottomrule
\end{tabularx}
\vspace{3pt}
\parbox{\linewidth}{\footnotesize\textit{Control:} Uniform also returns 322 in one call. This case illustrates evidence reuse and completion, rather than superiority over Uniform or a claim that the initial reasoning was entirely wrong.}
\end{table}

Table~\ref{tab:case} illustrates a quartic-root problem from the historical pool. None of the four candidates has an extractable final answer, but one contains a useful sum-of-squares identity. The moderator verifies that identity and completes the complex quadratic, obtaining 322. Uniform also succeeds, so the example explains evidence reuse rather than instruction superiority. The identity and root were checked independently. Absence of a selectable answer does not imply absence of useful intermediate computation; Appendix~\ref{app:cases} gives the provenance and a separate majority-error example.

\subsection{Token savings and sequential latency}
\label{sec:latency}
The F2 timing audit records no moderation time on agreement problems. Among 44 triggered problems, the added moderation time has median 108 seconds and 90th percentile 304 seconds; averaged across the full panel, the increment is 52 seconds per problem. Thirty-seven triggered problems use one round and seven use two. Sequential checking can therefore remain costly even when it produces fewer tokens than another batch of independent candidates.

The audit's recorded end-to-end ratio to SC@8 is 1.18 at the median and 1.41 at the 90th percentile on triggered problems. These are measurements of the logged serving configuration, which uses eager execution and single-sequence moderation. GPU sharing was not recorded, and generating only four proposals was not timed separately. The ratios are not a controlled estimate of a production four-proposal pipeline. We report token consumption as workload and retain latency as a deployment constraint, without translating the 46--47\% token reduction into a corresponding speedup.

\paragraph{What the controls establish.}
Under Hybrid scoring across two independently sampled pools, moderation improves on voting over the same four candidates with no observed regressions. The F2 answer-only intervention supports a contribution from visible derivations beyond answer counts alone. The F2 always-on control shows that checking agreement states adds computation without observed benefit. Together, these controls support reusing already-generated reasoning selectively. They do not establish that further independent sampling is less accurate, that the current route-specific instructions are necessary, or that fewer tokens imply lower latency.

\section{Related work}
\paragraph{Sampling and adaptive computation.}
Chain-of-thought prompting elicits intermediate reasoning~\citep{wei2022chain}; self-consistency aggregates sampled answers~\citep{wang2023selfconsistency}. Adaptive-Consistency adjusts sample counts using agreement~\citep{aggarwal2023adaptive}, while test-time compute allocation studies search and refinement as a function of difficulty~\citep{snell2024scaling}. Adaptive sampling and selective moderation both allocate further computation according to intermediate evidence. They differ in whether that computation generates independent attempts or inspects existing derivations.

\paragraph{Candidate selection and generative fusion.}
Universal Self-Consistency uses an LLM to select the most consistent candidate~\citep{chen2023universal}. LLM-Blender combines pairwise ranking with generative fusion of selected outputs~\citep{jiang2023blender}. Thus, both model-based selection and generation from candidates are established mechanisms. ReSolve studies what visible derivations and selective invocation add with shared candidates and fixed model weights. It also measures recovery and preservation failures relative to a shared fixed-pool selection oracle.

\paragraph{Verification and feedback.}
Process verification scores intermediate steps~\citep{lightman2023verify}; FOBAR reconstructs a masked problem value from a candidate answer~\citep{jiang2024forward}. Self-Refine revises an output through feedback~\citep{madaan2023selfrefine}. Multiagent debate and reconciliation exchange arguments~\citep{du2023multiagent,liang2024encouraging,chen2024reconcile}, and Mixture-of-Agents conditions later generations on preceding outputs~\citep{wang2024moa}. Our controls separate derivation access, invocation, and instruction choice within a shared candidate pool; Uniform's competitive results make these distinctions essential.

\paragraph{Search over reasoning states.}
Tree of Thoughts evaluates intermediate thoughts and backtracks~\citep{yao2023tree}; Graph of Thoughts permits richer dependencies~\citep{besta2024graph}. ReSolve begins with independent responses and uses their answer histogram to control bounded interventions. It neither maintains an explicit proof-state search structure nor assumes a step-wise reward model. No direct performance comparison with these search methods is claimed.

\section{Discussion: practical interpretation}
\label{sec:discussion}
The experiments suggest a simple allocation principle: once several expensive reasoning trajectories already exist, additional compute need not be spent only on independent resampling. When candidates disagree or lack a final answer, reading and extending existing derivations can recover useful work that voting discards. The answer-only control isolates the contribution of visible derivations under the shared output-cap rule, while the always-on control shows where additional verification was unnecessary on the evaluated pool. The comparison with \sceight{} also reveals complementary failure modes: further samples can supply a missing answer, while moderation can complete or correct reasoning already available.

For inference design, this motivates treating candidate generation and subsequent processing as separate decisions. A system can first inspect agreement, then read selected derivations when another call is warranted, while charging both reading and generation to its budget. ReSolve instantiates this pattern with the same existing model as proposer and moderator. The competitive Uniform control makes the practical lesson useful beyond one prompt family: candidate evidence and selective invocation are design choices that can be examined independently of specialized instructions. In this setting, already-generated reasoning is a resource for further computation, including when it has not yet produced a selectable answer.

\section{Conclusion}
Generating four solutions has already incurred the cost of their reasoning. ReSolve reuses that evidence and continues only when candidates disagree or lack an answer. Across two candidate pools, this improves four-candidate voting with no observed regressions and uses 46--47\% fewer tokens than eight-sample self-consistency, without establishing an accuracy advantage over the latter. Evidence removal and always-on checking support both reuse and selectivity under the tested protocol.

\label{maintextend}

\clearpage
\bibliographystyle{unsrtnat}
\bibliography{main}

\clearpage
\appendix
\small
\section{Implementation and inference protocols}
\label{app:implementation}
\paragraph{Answer extraction and grouping.}
The implementation first attempts to extract the last boxed answer. If none is available, it uses conservative short-answer patterns within the last 1,200 characters. Normalization removes presentation-level LaTeX delimiters and spacing, outer braces, and trailing periods, and lowercases the result. This grouping rule is not a symbolic-equivalence oracle. It is used consistently for invocation and voting, independently of the evaluation grader.

The controller first handles zero parseable answers, then a single answer group, then a strict majority among parseable answers, then a two-group split, and finally fragmented disagreement. For the fresh training-free protocol, evidence exposes the first representative of each answer group, its key, and its count, with at most six groups (two in the two-group state). Repeated support for an existing key changes its count without replacing the first representative; a new key can introduce its generated response as a representative. Unparseable responses are not shown when rendering answer groups. With no parseable answer, repair instead exposes up to the first six attempts in the current pool, including appended moderator attempts when they fit. Thus, feedback can update answer counts without exposing a revised argument. A representative uses at most 6,000 characters; long text retains a short head and an approximately 75\% tail with an explicit omitted-middle marker. The candidate block is capped at 36,000 characters and the rendered prompt at 48,000. A nonempty post-\texttt{</think>} derivation is preferred when available. These are evidence-construction limits; the fresh protocol applies no additional token-context clipping to an already-rendered moderator prompt.

\paragraph{Acceptance and fallback.}
At the beginning of a round, a single answer group commits without a new call. Otherwise the response is appended. A majority audit commits if it returns the current leading key; a two-group or fragmented intervention commits if it returns a key already present before that call. A generated key can also commit after appearing twice consecutively among parseable moderator responses; an unparseable response does not reset the stored streak. At the call limit, the fallback selects the most frequent parseable key and its latest supporting response. Ties follow the first encountered group order. If no answer can be extracted, the fallback is the first original response. The stored convergence label for consensus can occur after a call; invocation counts are therefore taken from call records rather than inferred from that label.

\paragraph{Principal-pool sampling.}
The proposer and moderator are Qwen3-4B-Thinking-2507, served with vLLM~\citep{kwon2023vllm}. Each problem receives eight separately sampled proposals. The first four proposals form the shared moderation pool; the additional four are used only by the eight-sample controls. Each call has output cap $32768-T^{\mathrm{in}}-32$, with at least 1,024 available output tokens required for a moderation call. Proposal temperature/top-$p$ are 0.7/0.9; moderation uses 0.1/1.0. F2 and F3 use these candidate-compaction and output-cap rules throughout. No forced-close rescue or reference-answer feedback is used in either pool.

\paragraph{Uniform moderation.}
Uniform shares the evidence, including counts, as well as the invocation and stopping rules. It replaces the specialized majority, two-group, and fragmented-disagreement instructions with a common instruction to verify and repair. The no-answer state retains the repair operation. Thus, it isolates an instruction intervention within the controller; it is neither count-blind nor an always-on verifier.

\paragraph{Historical studies.}
H1/D1 share a recorded 160-problem pool, adding 30 BRUMO25 questions, with a 16,384-token moderator output cap. D1 reuses relevant outputs and adds 102 agreement checks and 58 independent solves. Overlapping results are not independent replications, and these costs are not pooled with the fresh protocol.

\paragraph{F3 execution.}
F3 resamples the eight candidates for every problem while retaining the F2 problem texts and frozen inference, utility, and grading code. Candidate requests are batched across problems, and independent per-problem controller threads submit calls through a batching proxy. The final batched implementation was validated against sequential execution for equivalent per-problem controller behavior; debugging runs are excluded from the reported results. This execution change should not be interpreted as a controlled latency comparison with F2.

\clearpage
\section{Scores and paired comparisons}
\label{app:statistics}
\label{app:grading}
\begin{table}[htbp]
\centering\small
\caption{F2 per-task correct counts as Hybrid/CV. Task sizes are 30, 30, 40, and 30. Hybrid and CV score the same outputs; the main text uses Hybrid.}
\label{tab:f2tasks}
\begin{tabular}{cccccc}
\toprule
Method & AIME24 & AIME25 & CMIMC25 & HMMT25 & Total \\
\midrule
SC@4 & 26/26 & 24/24 & 25/24 & 16/16 & 91/90 \\
Oracle@4 & 27/27 & 25/25 & 27/26 & 20/20 & 99/98 \\
SC@8 & 27/27 & 25/25 & 26/25 & 16/16 & 94/93 \\
Oracle@8 & 27/27 & 25/25 & 28/27 & 20/20 & 100/99 \\
\method{} & 27/27 & 25/25 & 27/27 & 21/21 & 100/100 \\
Uniform & 27/27 & 26/26 & 25/25 & 19/19 & 97/97 \\
Direct32 & 23/23 & 24/24 & 20/19 & 15/15 & 82/81 \\
\bottomrule
\end{tabular}

\end{table}

\begin{table}[htbp]
\centering\small
\caption{Within-pool tests for ReSolve. Holm correction covers SC@8 and Uniform within each pool and score. SC@4 is a secondary comparison. Combined problem-level tests use a different analysis described in the text.}
\label{tab:statistics}
\begin{tabular}{cccccc}
\toprule
Pool & Score & Comparator & Fix/break & Exact $p$ & Holm $p$ \\
\midrule
F2 & Hybrid & SC@8 & 7/1 & 0.07031 & 0.14062 \\
F2 & Hybrid & Uniform & 4/1 & 0.37500 & 0.37500 \\
F2 & Hybrid & SC@4 & 9/0 & 0.00391 & --- \\
F3 & Hybrid & SC@8 & 7/4 & 0.54883 & 1.00000 \\
F3 & Hybrid & Uniform & 2/2 & 1.00000 & 1.00000 \\
F3 & Hybrid & SC@4 & 7/0 & 0.01562 & --- \\
F2 & CV & SC@8 & 8/1 & 0.03906 & 0.07812 \\
F2 & CV & Uniform & 4/1 & 0.37500 & 0.37500 \\
F2 & CV & SC@4 & 10/0 & 0.00195 & --- \\
F3 & CV & SC@8 & 7/4 & 0.54883 & 1.00000 \\
F3 & CV & Uniform & 2/2 & 1.00000 & 1.00000 \\
F3 & CV & SC@4 & 7/1 & 0.07031 & --- \\
\bottomrule
\end{tabular}

\end{table}

Hybrid is the main scoring view in this manuscript, consistent with the original evaluation protocols. CV supplies a sensitivity analysis of the same outputs. Under Hybrid, F2/F3 totals are 100/99 for \method{}, 94/96 for SC@8, and 97/99 for Uniform. Under CV, they are 100/97, 93/94, and 97/97. These are alternative scores of shared responses, not separate generations.

Within each pool, exact two-sided McNemar tests use fixes and breaks. The prespecified two-comparison family comprises SC@8 and Uniform; Holm correction is performed within each pool and scoring policy. The SC@4 comparison is secondary and was added after the initial evaluation. Its unadjusted $p$-values should not be read as prespecified evidence. Treating it as a third member of the F2 Hybrid Holm family would give $p=0.0117$, without changing the two original adjusted values.

The combined F2/F3 analysis groups repeated outcomes by problem and reports a sign-flip test of the average paired difference. It has 130 problem units with two candidate pools each. For SC@4 the Hybrid count difference is $16/260=6.15$ percentage points ($p=0.0003$); for SC@8 it is $9/260=3.46$ points ($p=0.093$); for Uniform it is $3/260=1.15$ points ($p=0.51$). The count denominator summarizes observations, whereas dependence is handled at the problem level in the test. These combined tests are reported from the completed experiment analysis; they are not obtained by applying McNemar to pooled outcomes as if every observation were independent. No adjustment across all exploratory analyses is claimed.

\subsection{Scorer disagreements and clipping}
\label{sec:grading}
\begin{table}[htbp]
\centering\small
\caption{Scoring sensitivity on shared outputs. F2 rule values are from the audit; F2 Hybrid/CV values are reconciled to archived outcomes. Provisional manual judgments do not replace recorded scores.}
\label{tab:scorers}
\begin{tabular}{cccccc}
\toprule
Method & F2 rule & F2 CV & F2 Hybrid & F3 CV & F3 Hybrid \\
\midrule
\method{} & 100 & 100 & 100 & 97 & 99 \\
Uniform & 97 & 97 & 97 & 97 & 99 \\
SC@8 & 94 & 93 & 94 & 94 & 96 \\
SC@4 & 91 & 90 & 91 & 91 & 92 \\
Direct32 & 81 & 81 & 82 & 79 & 80 \\
\bottomrule
\end{tabular}

\end{table}

In CMIMC25 index 16, SC@4 and SC@8 return $2271049/129600$, equivalent to the reference $1507^2/360^2$. The rule accepts the answer and CV rejects it. Thus F2's main SC@8 comparison is 100 versus 94 with 7/1 paired changes; the CV-only 100 versus 93 with 8/1 includes an equivalent-form false negative. Answer-only exhibits the same extra CV discrepancy. We retain each scoring policy's recorded result and do not count this disagreement as an additional mathematical repair.

F2 deduplicates candidate and returned-response texts into 1,258 unique scoring keys. The audit reports 15 rule/CV disagreements, including equivalent-form judgments and incomplete responses. The F2 ReSolve and Uniform outputs have no rule/CV disagreements. The report's provisional manual adjudications are not used as a third primary benchmark. This distinction avoids treating an analyst's judgment as a fully validated grading protocol.

To fit the verifier context, 291 scoring texts are clipped; the rule grader uses the full response. A replay examines every discordant output in the reported comparisons under both scoring views, covering 28 problems and 139 scoring inputs. Replayed flags match the recorded clipping flags. No clipping operation removes an existing final answer, and no ReSolve output in the SC@8 or Uniform comparison is clipped. Five extracted-answer sensitivity verdicts on the affected main-comparison questions agree with the original labels. These checks address answer loss under clipping without certifying the correctness of every retained derivation or verifier decision.
\begin{table}[htbp]
\centering\small
\caption{Reported replay of CV clipping on discordant F2 comparisons. Oracle comparisons inspect candidate scoring inputs as well as moderator outputs, so input counts can exceed problem counts. Rows overlap and must not be summed. No existing final answer is removed; clipped Answer-only outputs already lacked final answers before clipping.}
\label{tab:clippingaudit}
\begin{tabular}{cccc}
\toprule
CV contrast & Discordant problems & Clipped inputs & Final answers lost \\
\midrule
\method{} vs. SC@8 & 9 & 2 & 0 \\
\method{} vs. Uniform & 5 & 0 & 0 \\
\method{} vs. Answer-only & 8 & 2 & 0 \\
\method{} vs. Direct32 & 23 & 17 & 0 \\
\method{} vs. Oracle@4 & 6 & 17 & 0 \\
\method{} vs. Oracle@8 & 7 & 39 & 0 \\
\bottomrule
\end{tabular}

\end{table}

\clearpage
\section{Controlled evidence and invocation interventions}
\label{app:answeronly}
The answer-only intervention reuses the F2 questions and first four candidates. Each retained proposal block is reduced to its header, final answer, and vote count; no-answer blocks retain their attempt header and unparsed-answer marker. The same restriction applies to later moderator-generated candidates. The Full representation supplies the visible post-\texttt{</think>} written derivation where available, under the original renderer limits.

At each state, the output cap is $32768-T^{\mathrm{in}}_{\mathrm{Full,state}}-32$. The counterfactual Full prompt is tokenized but not sent to the model. After the two trajectories diverge, this matches the same state's Full representation, not the original run's later states. The controller, sampling settings, maximum of three calls, and patience of two are unchanged. First-round prompt hashes, token counts, and caps match the original Full run on all 44 triggered problems. Regrading those Full outputs reproduces all 44 original labels under both Hybrid and CV. The follow-up uses two single-sequence engines per GPU with cache utilization 0.42 rather than 0.88; it reports no other engine-setting change.

The seven Hybrid improvements over Answer-only are AIME24 index 25, CMIMC25 index 15, HMMT25 indices 15 and 5, AIME25 index 27, HMMT25 index 28, and CMIMC25 index 4. Four have parseable candidate answers; three have none. Full has no Hybrid loss. Under CV there is one additional recorded improvement, the equivalent-answer discrepancy described above. Of the 44 triggered answer-only problems, 29 use one call, nine use two, and six use three, yielding 65 calls. Twenty calls reach the output cap, including 15 on no-answer problems. Cost totals include the same proposal pool.

For always-on Uniform, the final-protocol control adds one generic verification call on each original agreement problem and reuses existing Uniform outputs elsewhere. An unparseable check falls back to the original response. The one-call and closed-loop variants share the same added agreement checks, so they are controls on the same underlying outputs rather than independent replications. Both show zero fixes and zero breaks among those 86 checks. Moderation token totals in the report are rounded, and percentages computed from them are approximate.

\clearpage
\section{Route and recovery details}
\label{app:routes}
\begin{table}[htbp]
\centering\small
\caption{F2 route-wise Hybrid correct counts, recomputed from archived per-question records. Agreement is defined by answer groups rather than correctness.}
\label{tab:f2routes}
\begin{tabular}{ccccccc}
\toprule
Initial state & $N$ & ReSolve & SC@4 & SC@8 & Uniform & Oracle@4 \\
\midrule
Agreement & 86 & 74 & 74 & 74 & 74 & 74 \\
Dominant answer & 18 & 14 & 12 & 12 & 11 & 14 \\
Two groups & 8 & 5 & 4 & 5 & 5 & 5 \\
Fragmented & 12 & 4 & 1 & 3 & 4 & 6 \\
No answer & 6 & 3 & 0 & 0 & 3 & 0 \\
\bottomrule
\end{tabular}

\end{table}

F2's initial route counts are 86 agreement, 18 dominant-answer, eight two-group, 12 fragmented, and six no-answer problems. Under Hybrid, the SC@4 fixes are 2, 1, 3, and 3 in the four triggered categories, with no breaks. Against SC@8 the corresponding outcomes are 2/0, 0/0, 2/1, and 3/0. Against Uniform they are 3/0, 0/0, 1/1, and 0/0. Both moderators solve three of six no-answer problems. These are descriptive strata rather than independent significance tests.

Table~\ref{tab:agreement} separates agreement by parseability. Only 54 of the 86 skipped problems have four parseable candidates; 14 have just one. Agreement is therefore consistency among available answers, not necessarily four-way agreement. The 74 SC@4-correct skipped problems and 17 SC@4-correct triggered problems account for all 91 correct votes. All 17 triggered correct votes are preserved; the nine fixes occur among the other 27 triggered problems. The archived F3 summary does not provide this conditional denominator or parseability breakdown.
\begin{table}[htbp]
\centering\small
\caption{F2 agreement states by the number of parseable initial candidates. SC@4 and ReSolve share the skipped outputs; all 86 added Uniform checks leave correctness unchanged. Counts are descriptive.}
\label{tab:agreement}
\begin{tabular}{ccc}
\toprule
Parseable candidates & Agreement problems & Correct under Hybrid \\
\midrule
1 & 14 & 6 \\
2 & 7 & 5 \\
3 & 11 & 10 \\
4 & 54 & 53 \\
\bottomrule
\end{tabular}

\end{table}

F2's three recoveries beyond Oracle@4 all arise in no-answer states; both losses occur in fragmented states. F3 reverses the net balance with two recoveries and three losses. Its ten no-answer states, compared with six in F2, also contain three of the four ReSolve losses to SC@8. The aggregate oracle is bounded only for selection from its specified pool and evaluator. It is neither a deployable baseline nor an upper bound on a generator with additional computation.

\clearpage
\section{Historical mechanism controls}
\label{app:historical}
\begin{table}[htbp]
\centering\small
\caption{H1 training-free results as Hybrid/CV counts ($N=160$). Task sizes are 30, 30, 30, 40, and 30. All rows share the same four-candidate pool.}
\label{tab:h1tasks}
\begin{tabular}{ccccccc}
\toprule
Method & AIME24 & AIME25 & BRUMO25 & CMIMC25 & HMMT25 & Total \\
\midrule
Oracle@4 & 27/26 & 25/25 & 24/24 & 30/30 & 20/20 & 126/125 \\
SC@4 & 23/23 & 24/24 & 24/24 & 29/29 & 19/19 & 119/119 \\
\method{} & 26/26 & 26/26 & 24/24 & 30/30 & 21/21 & 127/127 \\
Matched operation & 25/25 & 26/26 & 24/24 & 31/31 & 20/20 & 126/126 \\
Uniform, closed loop & 26/26 & 25/25 & 24/24 & 31/31 & 20/20 & 126/126 \\
Operation permutation 1 & 24/24 & 26/26 & 24/24 & 32/32 & 20/20 & 126/126 \\
Operation permutation 2 & 26/26 & 25/25 & 24/24 & 30/30 & 20/20 & 125/125 \\
\bottomrule
\end{tabular}

\end{table}

\begin{table}[htbp]
\centering\small
\caption{D1 on the H1 pool ($N=160$). Input/output columns count moderation only; total includes proposals. Calls are panel totals and tokens are per-question means. The original gate triggers on 58 questions.}
\label{tab:diagnostics}
\begin{tabular}{cccccc}
\toprule
Configuration & H/CV & Calls & Input/q (K) & Output/q (K) & Total/q (K) \\
\midrule
SC@4 & 119/119 & 0 & 0.00 & 0.00 & 87.67 \\
\method{} & 127/127 & 74 & 2.01 & 4.56 & 94.24 \\
Uniform, closed loop & 126/126 & 68 & 1.78 & 4.36 & 93.80 \\
One specialized call, original gate & 125/125 & 58 & 1.50 & 3.56 & 92.73 \\
One Uniform call, original gate & 126/126 & 58 & 1.50 & 3.74 & 92.90 \\
One Uniform call, validity gate & 126/126 & 86 & 1.71 & 5.82 & 95.20 \\
One Uniform call, always & 126/126 & 160 & 2.30 & 8.40 & 98.37 \\
\method{}, validity gate & 127/127 & 102 & 2.23 & 6.64 & 96.54 \\
Uniform loop, validity gate & 126/126 & 96 & 1.99 & 6.45 & 96.10 \\
Fresh solve, original gate (16K) & 119/118 & 58 & 0.06 & 5.73 & 93.46 \\
\bottomrule
\end{tabular}

\end{table}

H1 and D1 share a historical 160-problem candidate pool that additionally contains 30 BRUMO25 questions. Their overlapping rows are reused observations. All displayed historical configurations use existing model weights. H1's Hybrid oracle decomposition is four recoveries and three losses; the corresponding CV decomposition is five and three. Both satisfy the recovery/loss identity with their own oracle totals (126 and 125). On the ten no-answer problems, ReSolve solves four while SC@4 solves none. The 102 agreement states include nine errors, unchanged by D1's always-on Uniform.

In D1, one candidate-conditioned Uniform call and a fresh independent solve act on the same 58 triggered questions under a 16K output cap. Hybrid scores are 126 and 119; CV scores are 126 and 118. The original answer is used on skipped questions. This is supporting evidence about candidate-conditioned reasoning, but changing the prompt means it is less isolated than the final-protocol answer-only intervention. Selective and always-on Uniform are both 126 under Hybrid while consuming different moderation budgets. These historical costs are not pooled with the 32K-context F2/F3 protocols.

\clearpage
\section{Case-study evidence}
\label{app:cases}
The main-text quartic example and the geometry example below come from the historical panel. Their identifiers are zero-based cache indices, not official contest question numbers. Descriptions are edited mathematical summaries, not verbatim model transcripts.

\paragraph{CMIMC25, index 7.}
The polynomial is $P(x)=x^4+20x^3+29x^2-666x+2025$. Expanding $(x^2+10x-36)^2+(x+27)^2$ verifies the identity found in the fourth unfinished attempt and retained in the moderator's input. The first-quadrant root is $(-10+i+\sqrt{243+88i})/2$, giving the required sum 322. The identity, discriminant, and quadrant were checked independently. All four initial final answers are unparseable and marked incorrect under both scoring views. Their finish reasons are not recorded, so absence of an answer is not relabeled as observed length termination. The numeral 322 occurs in unrelated arithmetic in an initial trace; the claim concerns absence of a correct selectable final response. Uniform also succeeds, with 8,558 output tokens versus 7,502 for ReSolve.

\paragraph{AIME25, index 28.}
Three candidates answer 153 and one answers 104. The majority calculation omits an equilateral triangle. Checking the region decomposition gives $(208-55-49)\sqrt{3}=104\sqrt{3}$ and answer 104. ReSolve returns the correct answer after two calls and 21,421 output tokens. Uniform reaches 104 on its first call but returns to 153 on its second. This example concerns discovering and retaining a correction; it does not show that a generic verification instruction cannot find the correction.

\clearpage
\section{Evidence provenance and scope}
\label{app:provenance}
F2/H1/D1 counts and paired tests are recomputed from archived per-question records. F3, invocation, latency, evidence-removal, and clipping results use completed experiment reports; marginal and paired-count arithmetic is checked here, but remote generation and grading artifacts were not replayed locally. Combined problem-level tests are taken from the completed analysis. Historical protocols remain separate. The evidence concerns mathematical inference with the specified backbone and budgets.

\end{document}